\documentclass{article}

\usepackage[preprint]{nips_2018}

\usepackage[utf8]{inputenc} 
\usepackage[T1]{fontenc}    
\usepackage{hyperref}       
\usepackage{url}            
\usepackage{booktabs}       
\usepackage{amsfonts}       
\usepackage{nicefrac}       
\usepackage{microtype}      
\usepackage{graphicx}

\usepackage{graphicx}
\usepackage{amsmath}
\usepackage{mathrsfs}
\usepackage{appendix}

\usepackage{array}
\usepackage{multirow}
\usepackage{tabu}

\newtheorem{lemma}{Lemma}
\newtheorem{theorem}{Theorem}

\newtheorem{corollary}{Corollary}

\newtheorem{definition}{Definition}

\newenvironment{proof}{\paragraph{Proof:}}{\hfill$\square$}

\def\Nat{\mathbb{N}}
\def\Reals{\mathbb{R}} 
\def\T{\mathscr{T}}
\def\X{\mathscr{X}}
\def\diag{\mbox{diag}}
\def\cond{\mbox{$\chi$}}
\font\mathlet=eurm10 
\def\eff{\hbox{\mathlet f}}
\def\matrixw#1{J^#1}
\def\matrixx#1{K^#1}

\setcitestyle{numbers}

\title{Generalization in quasi-periodic environments}

\author{
  Giovanni Bellettini\\
  DIISM, University of Siena\\
  Siena, Italy\\
  \texttt{bellettini@diism.unisi.it}\\
  \And
  Alessandro Betti\\
  University of Florence\\
  Florence, Italy\\
  \texttt{alessandro.betti@unifi.it}\\
  \And
  Marco Gori\\
  SAILab, University of Siena\\
  Siena, Italy\\
  \texttt{marco@diism.unisi.it}\\
}

\begin{document}

\maketitle

\begin{abstract}
By and large the behavior of stochastic gradient is regarded as a 
challenging problem, and it is often presented in the framework of
statistical machine learning.
This paper offers a novel view on the analysis of on-line models of learning 
that arises when dealing with a generalized 
version of stochastic gradient that is based on dissipative dynamics. 
In order to face the complex evolution of 
these models, a systematic treatment is proposed which is based
on energy balance equations that are derived by means of the
Caldirola-Kanai (CK) Hamiltonian. According to these equations, learning 
can be regarded as an ordering process
which corresponds with the decrement of the loss function. 
Finally, the main results established in this paper is that 
in the case of quasi-periodic environments, where the pattern novelty
is progressively limited as time goes by, the system dynamics yields 
an asymptotically consistent solution in the weight space, 
that is the solution maps similar patterns to the same decision.   


\end{abstract}

\section{Introduction}
Stochastic gradient descent, also known as incremental gradient descent, is an approximation of the classic gradient descent optimization over a batch of data. 
In large scale experiments in machine learning, the optimization of the empirical 
risk computed according to classic optimization methods yields the batch mode
version of learning that, unlike stochastic gradient,  typically exhibits a remarkable computational burden. This has been clearly established in This is very effective in the case of large-scale machine learning problems~\cite{Bottou:2007:TLS,bottou2010large}.
However, the behavior of stochastic gradient is more
difficult to interpret then batch mode.  Most of the studies are carried out
in a statistical framework (see e.g.~\cite{Ferguson:1982:IML}).
The convergence of stochastic gradient descent has been studied
in the context of convex minimization and stochastic approximation. 
Basically, the application of the classic 
Robbins-Siegmund theorem~\cite{Robbins1971} leads to 
conclude that when the learning rates 
decrease with an appropriate rate, the stochastic gradient descent converges almost surely to a global minimum when the objective function is convex or pseudo-convex.
%
%
The analysis carried out in this papers holds for a more general class of 
learning systems that incorporate stochastic gradient as a special case. 
Instead of working within the framework of statistics we carry out an analysis
that relies aimed at capturing the environmental regularities in the framework 
of system dynamics that emerges  associated with the classic 
Caldirola-Kanai (CK) Hamiltonian~\cite{Herrera86}. It has been recently used to propose
an interpretation of learning according to the principle of 
least cognitive action~\cite{Betti:2016:PLC}, which gives rise to 
Euler-Lagrange differential equations for the synaptic connections. 
The most important novelty introduced in the paper arises from  the notion of
{\em quasi-periodic
environments}, which assumes that the input at a certain instant in time   
can asymptotically be mapped to an arbitrarily close pattern.
It is worth mentioning that this definition is inspired by the classic notion
of almost periodicity~\cite{Bohr1947}.
The hypothesis of quasi-periodicity plays an important role in capturing the essence of the 
generalization of the learning process. We show that when imposing 
the positiveness of the eigenvalues of the Jacobian of the gradient of the 
loss the quasi-periodicity hypothesis leads to conclude that on the 
convergence of the of a class of dynamical system that incorporate stochastic
gradient as a special case. The paper is organized as follows: In the next section
we introduce the cognitive action laws of learning while in
Section\ref{EnBal-sec}  we introduce the  corresponding energy balance.
In Section~\ref{Quasi-sec} we present the main results of the paper concerning
the convergence and in Section~\ref{Conc-sec} some conclusions are drawn.



\section{Cognitive action laws of learning}
\label{CAL-sec}
In most challenging and interesting learning tasks
taking place in humans, unlike machines, the underlying computational processes
does not seem to offer a neat distinction between the training and the test set. 
As time goes by, humans react surprisingly well to new stimuli, 
while keeping past acquired skills, which
seems to be hard to reach with nowadays intelligent agents.
This suggests us to look for alternative foundations of learning, which are not necessarily 
based on statistical models of the whole agent life.
We can think of learning  as the outcome of laws of nature 
that govern the interactions of intelligent agents with their own environment, regardless of their
nature. We reinforce the underlying principle  that the acquisition of cognitive skills by learning
obeys information-based laws on these interactions, which hold regardless of biology. 

%
%
%
The notion of time is ubiquitous in laws of nature. Surprisingly enough, most studies on
machine learning have relegated time to the related notion of 
iteration step. From one side the connection
is pertinent and apparently sound, since it involves the computational effort, that is also observed in nature. 
From the other side, notice that time acts as an index of any human perceptual input.
Now, while the iteration steps in machine learning somewhat parallel this idea, most
algorithms neglect the smoothness of the temporal information flow.  As a consequence,
we pass through the inputs of the training set that, however, turn out to be an
unrelated picture of artificial life, since the temporal relations are lost.  
 This might be one the main reasons of current
performance gap in challenging tasks of speech and vision understanding with respect
to humans. 

Here we discuss how to 
express and incorporate time in its truly continuous nature, so as the evolution of the weights 
of the neural synapses follows equations that resemble laws of physics.
%
%
The environmental interactions are modeled under the general framework of 
constraints on the learning tasks. 
In the simplest case of supervised learning, we are given the collection $\mathscr{L}=
\{(t_{\kappa},x(t_{\kappa})),y_{\kappa}\}_{\kappa \in \Nat}$ of
supervised pairs $(x(t_{\kappa}),y_{\kappa})$ over the temporal sequence 
$\{t_{\kappa}\}_{\kappa \in \Nat}$.
We assume that those data are learned by a feedforward neural network
characterized by the function 
$\eff(\cdot,\cdot): \Reals^{m} \times \Reals^{d} \rightarrow \Reals^{n},$
so as the input $x(t)$ is mapped to $y(t)=\eff(w(t),x(t))$.
Learning affects the synapses by changing the weights $w(t)$
during the agent life taking place in the horizon $\T=[0,T]$, 
where $T$ can become pretty large, so as the condition 
$T \rightarrow \infty$ might be reasonable.
%
%
\font\ninerm=cmr9
\font\eightrm=cmr8
\font\sixrm=cmr6
\font\manfnt=manfnt 
\def\becomes{\ifmmode\ \hbox\fi{\manfnt y}\ } 

\def\sdol{\global\catcode`\$=12 \global\catcode`\#=12}
\def\ndol{\global\catcode`\$=3  \global\catcode`\#=6}

\def\mixthree{\begingroup
  \def\\{\noalign{\penalty-200}}
    \halign\bgroup         
          \hfil$##$\hfil\quad&         
          \vtop{\noindent\hsize=11.9em ##\medskip}\cr}
          
\def\endmix{\egroup\endgroup}

%
%
The  environmental  interactions take place over 
a temporal manifold, so as the perceptual input space $\X \in \Reals^{d}$ is ``traversed''
by the map $\T \to \X \subset \Reals^{d}: \ t \to x(t)$. 
The following analysis holds for different types of constraints, including the
previous pointwise constraints of supervised learning where, each example $x_{\kappa}$
can be associated with the  loss
$V(t,w(t))=\Vert\eff(w(t),x(t))-y_{\kappa} \big\Vert^{2}  \cdot \delta(t-t_{\kappa})$.
When looking at the neural network we can regard $w\in\Reals^m$ as
 the {\em Lagrangian coordinates} of a virtual mechanical system, so as 
the potential  of the system is defined by over the Lagrangian coordinates $w$. 
In this perspective, as already noticed, we look for trajectories $w(t)$ 
that possibly lead to configurations with small potential energy. 
Following the duality with mechanics,  we also introduce the notion of kinetic energy.
Now, we parallel the notion of velocity, by considering how quickly the weights
of the connections are changing. We can also dualize the notion of
mass $m_{i}>0$ of a certain particle by introducing the
{\em mass} of a connection weight. In so doing, the overall system is 
characterized by the conjugate variables that correspond with
the {\em position} $w_{i}(t)$ and the {\em velocity} $\dot{w}_{i}$. 
Then, we define the {\em kinetic energy} as
\begin{eqnarray}
	K(\dot{w}) = \frac{1}{2} \sum_{i=1}^{m} m_{i} \dot{w}_{i}^{2}.
\label{KineticEnergyDef}
\end{eqnarray}
%
%
Now, let us consider a Lagrangian inspired from mechanics, which
is split into the potential $V(\cdot,\cdot)$ and the kinetic energy as 
\begin{equation}
	F_{m}(t,w,\dot{w}) = 
	\sum_{i=1}^{m} K(t,\dot{w}_{i}) - V(t,w).
\label{SL-Lagrangian}
\end{equation}

The Lagrangian $F_{m}$ has an intriguing meaning that arises
when we explore the underlying cognitive processes. Let us consider  
\begin{equation}
	F(t,w,\dot{w}):= \psi(t) \ 
	F_{m}(t,w,\dot{w})
\end{equation}
and define $\psi(t)$ as the {\em dissipation function} of the agent. 
The reason of the name will be captured later on when
discussing  energetic issues behind the emergence of learning.  
%
%
We are now ready to reformulate learning in the general framework of
environmental constraints. The living agent is characterized by a 
neural network whose  weight  $w(\cdot)$ vector is the one for which  
$\delta A \lvert=0$, where 
\begin{equation}
	A(w) = \frac{1}{T}\int_{0}^{T} F(t,w,\dot{w}) \ dt
\label{CoActGeEq}
\end{equation}
is the {\em cognitive action} of the system.
%
%
A comment on the optimization of the cognitive action~(\ref{CoActGeEq}) is in order. 
In general, the problem makes sense whenever the Lagrangian is given all over $[0,T]$.
The very nature of the problem results in the explicit time dependence of $F(\cdot,\cdot,\cdot)$, which, in turn, depends
on the information coming from the interactions of the agent with the environment. 
The underlying assumption in learning processes is that 
after the agent has inspected a certain amount of information, it will be able to 
make predictions on the future. 
Hence, whenever we deal with a truly {\em learning environment}, in which the agent
is expected to capture regularities in the inspected data, we can make the assumption that
the weight trajectory converges to an end-point that somewhat expresses the 
saturation of the agent learning capabilities. Hence, we make the 
following border assumptions:\index{day-night rhythm} 

\begin{align}
\label{LB-lin-cond}
	\lim_{t \rightarrow 0^+} \dot{w}_{i}(t) = 0, \qquad \lim_{t \rightarrow T^-} \dot{w}_{i}(t) = 0.
\end{align}
As it will be shown later, these conditions can be guaranteed if we assume
that long-life learning undergoes a {\em day-night rhythm} scheme. 
Such a scheme follows the corresponding human metaphor:
The perceptual information is only provided during the day, while the agent ``sleeps'' at night
without receiving any perceptual information, which is translated into the condition
$\dot{x}=0$. 
We assume to undergo a long-life learning scheme  which repeats days of life according
to the above rhythmic scheme. Before discussing this assumption, we start noticing that
in perceptual tasks, the day-night rhythm doesn't alter the semantics that can be captured from the
environmental information flow. Hence, just like an uninterrupted flow, this rhythmic interaction
keeps the semantics, but favors the simplicity and the effectiveness of learning processes, since 
the lack of night stimulus facilitates the 
verification of the condition~(\ref{LB-lin-cond}) on the right border. Unlike an uninterrupted flow,
the day-night rhythm allows us small weight updates from consecutive days. Hence, 
if $w(t_{\kappa})$ is the weight vector at the end of day $\kappa$, the day after,
the weight $w(t_{\kappa+1}) \simeq w(t_{\kappa})$, which means facilitates the
approximation of the condition $\dot{w}(t=t_{\kappa+2})=0$. Now,  let's
write the the equations of the weights $w_{i}(t)$ of the synaptic connections.
If we pose $D=d/dt$ be then the Euler-Lagrange equations 
$D F_{\dot w_{i}} - F_{w_i} = 0$ becomes
\begin{equation}
	m_{i} \ddot{w}_{i} + \frac{\dot{\psi}}{\psi} \dot{w}_{i} + V_{w_{i}}=0. 
\label{EL-eqs-CA}
\end{equation}
As we can see the dissipation function $\psi$, which is always positive, strongly
affects the neurodynamics.  In this case the above equation reduces to 
\begin{equation}
	m_{i} \ddot{w}_{i} + \frac{\dot{\psi}}{\psi} \dot{w}_{i} + V_{w_{i}}^{\prime}=
	m_{i} \ddot{w}_{i} + \theta \dot{w}_{i} + V_{w_{i}}=0,
\label{EL-eqs-DI}
\end{equation}
where the last reduction comes from choosing the dissipation function 
$\psi(t)=e^{\theta t}$. Notice that when $m_i\to 0$ the above differential
equation reproduces gradient descent, indeed in this limit
the Euler approximation of the resulting equation is $w(k+1)=w(k)-\eta V_w$.

\section{Energy balance}
\label{EnBal-sec}
Let $U(t):=V(t,w(t))+K(\dot{w}(t))$ be the {\em internal energy} of the agent
and let us define
\begin{equation}
	Z(T):=\sum_{i=1}^{m}  \int_{0}^{T}  \frac{\dot{\psi}}{\psi}  \dot{w}_{i}^{2} dt , \qquad 
	E(T):= \int_{0}^{T} \partial_{t} V(\tau,w(t)) dt.
\label{DevTermEnEq}
\end{equation}
For reasons that will become clear in the following, 
the term $Z(T)$ is referred to as the {\em dissipated energy}, while 
$E(T)$ is referred to as the {environmental energy}. The following theorem
states a fundamental invariance property of any learning agent based on
Eq.~(\ref{EL-eqs-CA}).
\begin{theorem}
The system dynamics described by Eq.~(\ref{EL-eqs-CA}) obeys the invariant
\begin{equation}
	Z + \Delta U - E = 0,
\label{EnBalWthDev}
\end{equation}
where $\Delta U=U(T)-U(0)$
\label{FirstEnergyInvariant}
\end{theorem}
\begin{proof}
From Eq.~(\ref{EL-eqs-CA}) we get
\begin{align*}
	\sum_{i=1}^{m} m_{i} \ddot{w}_{i} \dot{w}_{i} 
	+ \frac{\dot{\psi}}{\psi} \sum_{i=1}^{m}  \dot{w}_{i}^{2} 
	+ \sum_{i=1}^{m} V_{w_{i}}^{\prime} \dot{w}_{i} =0. 
\end{align*}
Now, we have $D K(\dot{w}(t))=D \big(1/2 \sum_{i=1}^{m}
m_{i} \dot{w}_{i}^{2} \big) = \sum_{i=1}^{m} m_{i}\ddot{w}_{i} \dot{w}_{i}$
and 
\[
	D V(t,w(t)) - \partial_{t} V(t,w(t))= \sum_{i=1}^{m} V_{w_{i}}^{\prime} \dot{w}_{i}.
\]
If we plug these identities into the above balancing equation
and integrate over the domain $[0,T]$ we get
\begin{align*}
	\int_{0}^{T} DK + \sum_{i=1}^{m} \int_{0}^{T} \frac{\dot{\psi}}
	{\psi} \dot{w}_{i}^{2} dt + \int_{0}^{T} DV - \int_{0}^{T}\partial_{t} V(t,w(t)) dt = 0
\end{align*}
from which the thesis immediately follows.
\\
\end{proof}\\
This balance offers a deep interpretation of the learning process.
The internal energy $U$ reduces to the potential energy in case
of stationary points with no kinetic energy. Clearly, its reduction 
$\Delta U <0$ is the sign of a learning process that has produced
an ordered configuration. When choosing a monotonic
non-decreasing developmental function $\psi$ then 
$\dot{\psi}/\psi \geq 0$ and, if $\psi$ is not  constant
then the energy term $Z$ is positive. This represents the energy that is dissipated
during the learning process in the interval $[0,t]$. In order to grasp the 
meaning of the energy term $E$, suppose we begin for $t=0$ from
a configuration defined by weights $w = \bar{w}$ such that for the subsequent input 
$x(t), \ t \in (0,T]$ we have $V(t,\bar{w})=0$. Clearly,
\begin{equation}
	\frac{d V}{d t} = \left[\frac{\partial V}{\partial w} \dot{w}\right]_{w=\bar{w}}  
	+ \frac{\partial V}{\partial t} = \frac{\partial V}{\partial t} =0,
\label{PerfectLearning}
\end{equation}
and, therefore, $E = 0$. On the opposite, suppose that beginning from 
the previous configuration, the neural network is fed with an input $x(t)$
in $[0,T]$ such that  $V(T,\bar{w}) = {\cal V}(x(T),\bar{w}) >0$.
Basically, in this case, the input $x(t)$, yields an error with respect to the 
chosen loss function. It turns out that the configuration defined by 
the weight vector $\bar{w}$ does not incorporate the additional information 
coming from the environment yet. For this reason, $E$ is referred to
as {\em environmental information}. Interestingly, $E$ captures the
generalization quality of the underlying learning process, so as 
when $E$ approaches the null value we are definitely in front of
good generalization. When looking at the variations of $V$, while 
$\partial_{w }V$ expresses the synaptic changes of the neural network during the
learning, the term $\partial_{t }V$, integrated over $[0,T]$ is a nice index
of the progress of generalization. 

From Theorem~\ref{FirstEnergyInvariant} we immediately end up into the following
corollary.
\begin{corollary}
The variation of the internal energy is bounded by the environmental energy
$
	\Delta U \leq E.
$
\label{BoundedDU}
\end{corollary}
This statement is especially interesting in the case in which the energy balance is
applied when $K(0)=K(T)=0$, since in this case the variation of the internal energy
becomes $\Delta U = V(T)-V(0)$.

\section{Quasi-periodic environments}
\label{Quasi-sec}
According to the previous discussion on the environmental energy, 
the  learning of the environmental information over the interval $[0,T_L]$ with $T_L<T$
requires the satisfaction of Eq.~(\ref{PerfectLearning}) in $[0,T_L]$.
A possible way of achieving this stationary condition is that of processing over
an environment created by repeating  the temporal segment $[0,T_L]$.
Clearly this condition corresponds with stating that the environmental energy
over $[0,T_L]$ is null and, moreover, this enables us to associate the environmental
energy on any temporal segment to the  {\em accuracy}. Clearly, the two notions
only matches in the case in which ${\cal V}(x(t),\bar{w})=0$, where ${\cal V}(x(t), w(t)):=V(t,w(t))$. 
As usual, the satisfaction of the above learning condition does not necessarily 
indicate an appropriate learning process, since one must also know the agent behavior
when generalizing to new consistent information. We propose to characterize
this consistency by introducing a notion that is very related to typical
statistical machine learning assumptions. Regardless of the specific probability distribution
that characterizes the environmental data, when focussing on $x(t)$ 
one can reasonable expect that a similar pattern had
already appeared in the past. Moreover, it makes sense to assume that
such a property take place uniformly in the temporal domain, and that
for any $t \in [0,T]$, at least one pattern in the past, at time $\overline{t}$,
be similar in the metric sense, that is $\lVert x(t) - x(\bar{t})\lVert$ is small.   
Of course, as $T \rightarrow \infty$ one can also reasonably expect that 
$\lVert x(t) - x(\bar{t})\lVert \rightarrow 0$. Let us consider the framework
dictated by the following specific asymptotical definition.
\begin{definition}
An environment is {\em quasi-periodic} in $[0,T]$ of order $p\in\Reals$ if there exist 
$\alpha>0$, $\epsilon>0$  and a positive differentiable
function $\tau: [0,T] \to (0,\infty]$ with $\tau(t)>0$
such that $\gamma(t):=t+\tau(t)$, satisfies
$\gamma'(t)>1$ and 
\begin{equation}
  \forall t\in [0,T] \quad \hbox{we have}\quad
  \lVert x(t) - x(\gamma(t)) \lVert \leq \frac{\epsilon}{(\alpha+t)^{p}}.
\label{Quasi-Periodic-Def}
\end{equation}
\end{definition}
In the simplest case in which $\tau(t) \equiv \tau$ the above definition 
reduces to an extended  notion of periodicity in which we are ``tolerant''
with respect to the match induced by the period. In general it is interesting to 
pick up $\tau$ from functional spaces equipped with classic analytic properties.
%
Now, we will show that the assumption on quasi-periodic environments has important
consequences on the learning process. In particular, we start studying its 
asymptotic behavior by analyzing the system dynamics defined by Eq.~(\ref{EL-eqs-DI})


Given two points vectors $x(t)$ and $w(t)$ we define the matrices $(J^w(\xi(t)))_{ij}= \partial_{w_j} {\cal V}_{w_i}(\eta_i(t),
\zeta_i(t))$ and  $(J^x(\xi(t)))_{ij}=\partial_{x_j} {\cal V}_{w_i}(\eta_i(t),
\zeta_i(t))$ where $\xi_i(t)=(\eta(t),\zeta_i(t))\in \Reals^{d+m}$. Similarly for ${\cal V}_x$ we can define the corrisponding
matrices $\matrixx{w}(\xi(t))$ and $\matrixx{x}(\xi(t))$

\begin{theorem}
	Let us consider the dynamical system described by Eq.~(\ref{EL-eqs-DI})
	and assume that the following conditions hold true 
	\begin{enumerate}
	\item [i.] {\em positiveness of} the spectrum of  $\matrixw{w}$: $\lambda_{m} = \min_{i} \inf_{\lambda \in [0,1]}
          \lambda_{i}(\xi(t)) > 0$ for all point $\xi(t)$ that lies in the segment joining
 $(x(t), w(t))$ and $(x(\gamma(t)), w(\gamma(t)))$
		where $\lambda_{i}(\xi(t))$, with $i=1,\ldots,n$, is one of the $n$
		eigenvalues of $\matrixw{w}$.
	\item [ii.] {\em Quasi-periodicity}: The environment is quasi-periodic of order $p>0$, $p\ne 1/2$.
	\end{enumerate}
	Then there exists $B_{w}>0$ such that the 
	weight variation over the pseudo-period $\tau(t)$ is bounded by
	\begin{equation}
		\lVert w(t)-w(\gamma(t)) \rVert \leq \frac{B_{w}}{(\alpha+t)^{p-1/2}}.
	\end{equation}	
\label{HomoExpConv}
\end{theorem}
\begin{proof}
Let us consider the dynamics at $t$ and $\gamma(t)=t-\tau(t)$:
\begin{align}
\begin{split}
	&\ddot w(t) + \theta \dot w(t) +  {\cal V}_{w}(x(t),w(t))=0\\
	&\ddot w(\gamma(t)) + \theta \dot w(\gamma(t))+  
	{\cal V}_{w}(x(\gamma(t)),w(\gamma(t)))=0.
\end{split}
\label{TwoEqDiff}
\end{align}
Now, if we pose $\omega(t):=w(t)-w(\gamma(t))$, from %
the mean value theorem, we know that for each $t\in [0,T]$ there exists a point $\xi(t)$ in $\Reals^{d+m}$
in the segment joining $(x(\gamma(t)), w(\gamma(t))$ and $(x(t), w(t))$ such that
\begin{align*}	
	{\cal V}_{w}(x(t),w(t))-{\cal V}_{w}(x(\gamma(t)),w(\gamma(t)))
  =\matrixw{w}(\xi(t))\omega(t) +\matrixw{x}(\xi(t))(x(t)-x(\gamma(t))) .
\end{align*}
We shall assume that $t\mapsto \xi(t)$ sufficiently regular. Then, from Eq.~(\ref{TwoEqDiff}) we get 
\begin{equation}
  \ddot \omega(t) + \theta \dot \omega(t) 
  +\matrixw{w}(\xi(t))\omega(t)+\matrixw{x}(\xi(t))(x(\gamma(t))-x(t))=0.
\label{GlobalEqwithFT}
\end{equation}
Because of the assumption on the positiveness of the spectrum of  $\matrixw{w}(\xi(t))$, 
from Lemma~\ref{HomoLemma}, we know that the associated homogeneous
equation is exponentially stable. 
Now we can reduce Eq.~(\ref{GlobalEqwithFT}) to the first order by setting 
$z_{1}=\omega$ and $z_{2}=\dot{\omega}$, so as we reduce to  
\begin{equation}
	\dot{z}= \begin{pmatrix} 0 & I\\
	-\matrixw{w}(\xi(t))  & -\theta I 
	 \end{pmatrix} z + 
	 \begin{pmatrix} 
	 	0 \\  -\matrixw{x}(\xi(t))(x(t)-x(\tau(t)))
	 \end{pmatrix}
\end{equation}
Now, since the environment is quasi-periodic
\[
	\left \lVert \matrixw{x}(\xi(t))(x(t)-x(\tau(t))) \right\rVert
	\leq \left\lVert  \matrixw{w}(\xi(t))\right\rVert 
	\cdot \lVert x(t)-x(\tau(t)) \rVert
	\leq B_{J} \frac{\epsilon}{(\alpha+t)^{p}}
\]
Finally, we can apply Lemma~\ref{FromExpSta2BIBO}, from which
the conclusion is drawn. 
\end{proof}

\begin{theorem}
 	Let us consider an agent working on an environment $\mathscr{E}$ under the following 
	assumptions
	\begin{enumerate} 
	\item [i.] There exists $C_{x} \in \Reals$ such that 
		$\sup_{t\ge 0}\lVert \dot{x} \lVert=C_x<\infty$ 
	\item	[ii.] The  matrices $K^w$ and $K^x$ are bounded.
        \item [iii.] $\mathscr{E}$ is a quasi-periodic environment of order $p>3/2$.
	\end{enumerate}
	Then its environmental energy is bounded on $(0,+\infty)$:
	\begin{equation}
		\left|\int_{0}^{\infty} {\cal V}_x(x(t),w(t))\cdot \dot x\, dt\right|\le C_E.
	\label{AsymptoticEEZero}
	\end{equation}
\label{GeneralizationTh}
\end{theorem}
\begin{proof}
  Define $\varphi(t):=\Vert{\cal V}_x(x(t), w(t))\Vert$.
  From the mean value theorem on ${\cal V}_x$ and Theorem~\ref{HomoExpConv} we have
  \[\Vert{\cal V}_x(x(t),w(t))\Vert\le
    \Vert{\cal V}_x(x(\gamma(t)),w(\gamma(t)))\Vert+
    \frac{B_w}{(\alpha +t)^{p-1/2}},\]
  that is $\varphi(t)\le \varphi(\gamma(t))+B_w/(\alpha +t)^{p-1/2}$. We now integrate both sides
  in $(0,T)$ we get $\int_0^T \varphi(t)\, dt\le \int_0^T \varphi(\gamma(t))\, dt+\int_0^T B_w/(\alpha +t)^{p-1/2}\, dt$.
  Now since
  \[\int_0^T \varphi(\gamma(t))\, dt=\int_{\gamma(0)}^{\gamma(T)}\frac{\varphi(s) ds}{\gamma'(\gamma^{-1}(s))}\le
    \frac{1}{\gamma'(\gamma^{-1}(0))}\int_{\gamma(0)}^{\gamma(T)}\varphi(s)\, ds\le
    \frac{1}{\gamma'(\gamma^{-1}(0))}\int_{0}^{\gamma(T)}\varphi(s)\, ds.\]
  Then we can conclude that
  \[\int_{0}^{T}\varphi(s)\, ds-\frac{1}{\gamma'(\gamma^{-1}(0))}\int_{0}^{\gamma(T)}\varphi(s)\, ds
    \le\frac{2 B_w}{2p -3}\left(\frac{1}{\alpha^{p-3/2}}-\frac{1}{(\alpha+T)^{p-3/2}}\right).\]
  Now taking the limit $T\to\infty$
  \[\int_{0}^{\infty}\varphi(s)\, ds\le \frac{\gamma'(\gamma^{-1}(0))}{\gamma'(\gamma^{-1}(0))-1}\frac{2 B_w}{(2p -3)\alpha^{p-3/2}}.\]
  Then we can conclude that
  \[
    \left|\int_{0}^{\infty} {\cal V}_x(x(t),w(t))\cdot \dot x\, dt\right|\le C_x\int_0^\infty\varphi(s)\, ds\le
    \frac{\gamma'(\gamma^{-1}(0))}{\gamma'(\gamma^{-1}(0))-1}\frac{2 B_w C_x}{(2p -3)\alpha^{p-3/2}}.
  \]
\end{proof}

      \begin{theorem}
        Under the assumptions of Theorem~\ref{GeneralizationTh}
        the dynamics of system~(\ref{EL-eqs-DI}) is convergent, that is
	there exists $\overline{w}$ such that $\lim_{t \rightarrow \infty} w(t) = \overline{w}$.
      \end{theorem}
      \begin{proof}
        The proof is gained by contradiction. Suppose there is no
	convergence to $\bar{w}$. Since $\Delta U$ is bounded, then
	 if we apply Theorem~\ref{GeneralizationTh}
	 we immediately end up into contradiction.
        \end{proof}

%
%
%


\section{Conclusions}
\label{Conc-sec}
In this paper we have introduced a new framework for understanding the convergence of a wide class of learning systems, which also includes stochastic gradient descent. The given result is mostly based on the introduction of the concept of quasi-periodicity, that seems to parallels classic statistical assumptions. The proof of the convergence arises from the expression of a bound on the environmental energy, that is also shown to be connected with the performance of the learning system. Hence, the given bound is a good measure of the system generalization capabilities.


\appendix
\section{Stability results on time-varying linear systems}
%
%
Suppose $x(t) \in \Reals^{n}$ and $A(t), B(t)$ are smooth
$n \times n$ matrices.  The dynamics of 
the  system 
\begin{equation}
	\ddot{w} + 2 A(t) \dot{w} + B(t) w=0
\label{LIN-VAR-AsyStab}
\end{equation}
exhibits nice stability properties under opportune assumptions
on matrices $A(t)$ and $B(t)$. 
In particular, we are interested in exponential stability, that is systems
for which there exists positive constants $\lambda$ and $\gamma$ such that
for all real $t$ and $t_0$ with $t>t_0$
\[\Vert\Phi(t,t_0)\Vert\le \gamma e^{-\lambda(t-t_0)},\]
where $\Phi$ is the transition matrix of the system.
In order to state an important 
stability property we need to introduce the matrix measure
\cite{Desoer1975} 
\begin{equation}
	\mu(P) = \lim_{h \rightarrow 0^{+}}
	\frac{\lVert I + h P\rVert-1}{h}
\label{MatrixMeasureDesoer}
\end{equation}
induced by the matrix norm $\lVert P\rVert$.
	
\begin{lemma} (Sun et al, 2007)\\
	The dynamical system defined by Eq.~(\ref{LIN-VAR-AsyStab})
	is exponentially stable if there exists
	a positive constant $m$ such that 
	\begin{equation}
		l+\sqrt{l^{2}+4 c} - 2 m < 0
	\label{PsuedoQ-eq}
	\end{equation}
	where $l=\sup_{t \geq 0} \max[0,2\mu(mI-A(t))], \ \ 
	c=\sup_{t \geq 0} \lVert 2 m A(t) - m^{2}I - B(t) \rVert$.
	Moreover, the condition $\lVert w(t)\rVert \leq M \exp(-\lambda t)$
	is verified for 
	$
		0<\lambda \leq m- \big(l+\sqrt{l^2+4v} \big)/2
	$.
\label{Exp-StabilityLemma}
\end{lemma}
\begin{proof}
See~\cite{DBLP:journals/ijcon/SunWZ07}.
\end{proof}

\begin{lemma}
Let us consider the homogeneous differential equation
\begin{equation}
	\ddot\omega(t) + \theta \dot\omega(t) + 
	B(t)\omega(t)=0.
\label{HomoAssEq}
\end{equation}
If the spectrum of $B(t)$ is positive and if $\theta^2\ge 4 \lambda_{\textrm{min}}$,
$\theta^2\ge 4\lambda_{\textrm{min}}\chi(1+\chi)/\chi$ for suitable constants $\lambda_{\textrm{min}}$ and
$\chi$,
then the dynamical system defined by Eq~(\ref{HomoAssEq}) is exponentially stable.
\label{HomoLemma}
\end{lemma}
\begin{proof}
Once we pose $A(t) = \frac{1}{2}\theta I$ 
we can apply Lemma~\ref{Exp-StabilityLemma}.
We start noticing that if we  choose $m<\theta/2$ then, 
from Eq.~(\ref{MatrixMeasureDesoer}), we get 
$\mu(mI-A(t))=\mu((m-\frac{\theta}{2})I)<0$ and
$\max[0,2\mu((m-\frac{\theta}{2})I)]=0$, so as
$l=\sup_{t \geq 0} \max[0,2\mu((m-\frac{\theta}{2})I)]=0.$
As a consequence, for the condition~(\ref{PsuedoQ-eq}) to be 
verified we need to satisfy $\sqrt{c} < m$. 
Now let us consider $c=\sup_{t \geq 0} \lVert  (m \theta - m^{2})I - B(t) \rVert$.
Since $J_{{\cal V}_{w}}(t)$ has positive eigenvalues, there exists $P(t)$ such that
 $B(t) = P(t) \Lambda(t) P^{-1}(t)$, where $\Lambda(t) = \diag(\lambda_{i}(t))$.
 Hence, the previous condition becomes
 \begin{align*}
 	c&=\sup_{t \geq 0} \lVert  (m \theta - m^{2})I - B(t) \rVert
	= \sup_{t \geq 0} \lVert P(t) \big((m \theta - m^{2})I -  \Lambda(t) \big) P^{-1}(t)\rVert\\
	&\leq \sup_{t \geq 0} \ \left\lVert \big((m \theta - m^{2})I -  \Lambda(t) \big) \right\lVert 
	 \sup_{t \geq 0} \big(\lVert P(t) \rVert \  \lVert P^{-1}(t)\rVert \big)\\
	&\leq |m \theta - m^{2} - \lambda_{m}| \cond < m^{2},
\end{align*}
where $\chi=\sup_{t \geq 0} \lVert P(t) \rVert \cdot \lVert P^{-1}(t)\rVert$
and $\lambda_{\textrm{min}} = \min_{i} \inf_{t \geq 0} \lambda_i(t)$. Now
take $m \theta - m^{2} - \lambda_{m}\ge0$, that is to say $\theta/2-1/2\sqrt{\theta^2-4\lambda_{\textrm{min}}}\le m\le \theta/2+
1/2\sqrt{\theta^2-4\lambda_{\textrm{min}}}$ The above inequality is satisfied for
\[
	m > \frac{\theta \chi - \sqrt{\chi^{2} \theta^{2} - 4 \chi(1+\chi)\lambda_{\textrm{min}}}}
	{2 (1+\chi)}.
\]
From positiveness of $B(t)$ 
we know that $\lambda_{\textrm{min}}>0$ and, therefore, the above bound yields for $m>0$. Moreover, all these bounds 
are coherent with the above requirement $m<\theta/2$. We only need to control the positiveness of the
square roots: we need $\theta^2\ge 4 \lambda_{\textrm{min}}$ and $\theta^2\ge 4\lambda_{\textrm{min}}\chi(1+\chi)/\chi$.
However once $\theta$ is fixed one can always satisfy these conditions by a rescaling of $B(t)$.
Finally, we conclude on the exponential stability
of differential equation~(\ref{HomoAssEq}).
\end{proof}

Based on classic results on the links between exponential and BIBO stability 
(see~\cite{Brockett2015}, Theorem 1, pag. 196), the following lemma establishes a 
sharper property on the asymptotical relation between the input and the output of
an exponentially stable system.
\begin{lemma}
Let $A$ be bounded on $(0,+\infty)$. Let us assume that the autonomous system
\begin{equation}
  \dot{\omega}(t) = A(t) \omega(t) + u(t)
\label{GenTVDiffEq}
\end{equation}
associated with the above differential equation is exponentially stable 
and that $\lVert u(t)\rVert = O(t^{q})$ with $q<0$, $q\ne -1/2$. 
Then $\lVert \omega(t)\rVert = O(t^{q+1/2})$.
\label{FromExpSta2BIBO}
\end{lemma}
\begin{proof}
        For any $t\ge t_0$ the solution of Eq.~(\ref{GenTVDiffEq}) can be
	expressed by
	\[
		\omega(t) = \Phi(t,t_0) \omega(t_0) + \int_{t_0}^{t} \Phi(t,\beta) u(\beta) d\beta.
	\]
	Since the autonomous system is asymptotically stable, when taking the norm of both sides we get
	\begin{align*}
		\lVert \omega(t) \rVert &\leq 
		\left\lVert \Phi(t,t_0) \omega(t_0) \right\rVert+ 
		\int_{t_0}^{t} \left\lVert \Phi(t,\beta) u(\beta) \right\rVert d\beta
		\leq \ M e^{-\lambda(t-t_0)}		
		+ \int_{t_0}^{t} \left\lVert \Phi(t,\beta) \right\rVert \cdot \left\lVert  u(\beta) \right\rVert d\beta\\
		&\leq  M e^{-\lambda(t-t_0)}+ 
		\left(\int_{t_0}^{t} \left\lVert \Phi(t,\beta) \right\rVert^2 d\beta\right)^{1/2}
		\left(\int_{t_0}^{t} \left\lVert  u(\beta) \right\rVert^2 d\beta\right)^{1/2}\\
		&\leq M e^{-\lambda(t-t_0)} + B_{u}\gamma
		\left(\int_{t_0}^{t} e^{-2\lambda(t-\beta)}\, d\beta\right)^{1/2}
		\left(\int_{t_0}^{t} \beta^{2q}\, d\beta\right)^{1/2} \\
		&=  M e^{-\lambda(t-t_0)} + 
		\frac{\gamma B_{u}}{\sqrt{2\lambda(2q+1)}}\bigg(1-e^{-2\lambda(t-t_0)}\bigg)
		\bigg(t^{2q+1}-t_0^{2q+1} \bigg)^{1/2}.
	\end{align*}
	Now, if we set $t = \mu t_0$, with $\mu>1$ we get
	\begin{align*}
		\lVert \omega(t) \rVert &\leq M e^{-\lambda \frac{\mu-1}{\mu}t} + 
		\frac{\gamma B_{u}}{\sqrt{2\lambda(2q+1)}}\bigg(1- e^{-2\lambda \frac{\mu-1}{\mu}t}\bigg)
		\bigg(1-\frac{1}{\mu^{2q+1}}\bigg)^{1/2}t^{q+1/2} = O(t^{q+1/2}).
	\end{align*}
\end{proof}



\bibliography{nn}
\bibliographystyle{unsrt}
\end{document}